\documentclass[11pt]{article}
\usepackage[margin=1in]{geometry}
\usepackage{amsmath}
\usepackage{amssymb}
\usepackage{booktabs}
\usepackage{graphicx}
\usepackage{hyperref}
\usepackage{natbib}
\usepackage{xcolor}
\usepackage{placeins}
\usepackage{float}
\usepackage{microtype}
\hypersetup{colorlinks=true, linkcolor=blue, citecolor=blue, urlcolor=blue}

\title{\textbf{CASCADE: An Agentic Regulatory Network Framework for Patient-Data-Validated Downstream Perturbation Prediction}}
\author{Jose A. Bird \\ \small ORCID: 0009-0006-2744-0606 \\ \small Affiliation: Independent Researcher \\ \small Correspondence: jbird@birdaisolutions.com}
\date{}

\begin{document}
\maketitle

\begin{abstract}
CASCADE is an agentic framework that predicts downstream transcriptional effects of gene perturbation from precomputed ARACNe regulatory networks, exposed through a LangGraph-orchestrated workflow and the Model Context Protocol (MCP). Because this perturbation-prediction capability is CASCADE's central function -- the specific claim an MCP-exposed tool call would surface to a downstream agent or user -- we introduce a patient-data validation methodology that tests whether that claim is trustworthy, not merely plausible. Every experiment in this paper invokes \texttt{Cascade\allowbreak Workflow\allowbreak.run()} directly with \texttt{analysis\_depth="focused"}, CASCADE's real, public agentic entry point. We confirm empirically, not merely by reading the routing code, exactly which internal analyses this triggers for the gene roles tested here (Section~\ref{sec:architecture}); only the resulting perturbation-effects output is used in this paper's statistical tests.

CASCADE extends the multi-agent orchestration pattern introduced by RegNetAgents \citep{regnetagents} to a complementary analytical direction. Rather than classifying a focal gene's \emph{upstream} regulators by cross-network source, CASCADE predicts the \emph{downstream} transcriptional consequences of perturbing a focal gene via directed propagation through tumor-specific ARACNe networks. Validating downstream predictions is harder than validating regulator-candidate lists, because it requires either sparse experimental perturbation data or a genotype-driven natural-experiment proxy. We use the latter -- focal-gene copy-number amplification as a dosage-based proxy for the inverse of knockdown -- tested against real patient tumor expression data rather than curated cancer-gene annotation lists.

For MYC, CASCADE's predicted knockdown targets show strong directional concordance with real patient expression differences between amplified and non-amplified tumors, across three TCGA cancer types (BRCA: 90.0\%, $p<10^{-8}$; COAD: 72.0\%, $p=0.0013$; STAD: 85.7\%, $p<0.0001$). Permutation baselines confirm these results are not artifacts of gene-set structure, landing within four points of the theoretical 50\% chance level in every case. The result survives a PAM50 breast-cancer-subtype control (92\% concordance within Luminal B alone), and it replicates in METABRIC, an independent cohort with no data overlap with CASCADE's network-construction data (87.2\%, $p<10^{-6}$). We also compared CASCADE against two curated, identity-matched gene-set baselines using Fisher's exact test (Section~\ref{sec:baseline_comparison}). An exhaustive curated MYC-target gene set scores within 1--4 percentage points of CASCADE across all three cancer types (ahead in BRCA and STAD, behind in COAD), and CASCADE's E2F3 rate is nominally higher than its curated comparison (96.0\% vs.\ 90.7\%). Neither difference reaches statistical significance ($p=0.34$--$0.86$ and $p=0.384$ respectively). CASCADE's own direction-calling clearly matters -- forcing its predictions to a uniform guess drops concordance by 14--29 points in every panel tested -- but its overall accuracy is not shown to exceed existing public knowledge of MYC- or E2F-driven biology (Section~\ref{sec:baseline_comparison}).

Beyond MYC, validation is gene-specific rather than universal. Seven further proliferation-machinery-associated regulators (E2F3, CCND1, AURKA, CCNE1, FOXM1, TOP2A, RPS6KB1) replicate the pattern cleanly in BRCA (89.6--100\% each), and two more (MDM2: 68\%, CCND3: 76\%) replicate more weakly. Four lineage-identity transcription factors (SOX9, FOXA1, GATA3, ESR1: 38.8\%, 34.0\%, 0.0\%, 4.0\%), a receptor tyrosine kinase (ERBB2, 18.0\% in BRCA, further inconsistent across cancer types below), and CCND2 do not. CCND2 is a cyclin-D paralog of the two CCND genes that do validate (CCND1, CCND3), and fails outright in the same cancer type (BRCA) where both paralogs succeed -- the clearest single-gene counterexample to a clean proliferation-machinery/lineage-identity rule. ESR1 was selected specifically because the proliferation-machinery/lineage-identity hypothesis below predicted it would fail, which it did. Extending seven of these proliferation-machinery genes to a second cancer type, where amplification frequency allowed it, shows the split is not purely gene-class-determined. AURKA (COAD, 92.0\%), CCNE1 (STAD, 98.0\%), TOP2A (STAD, 100.0\%), and CCND3 (STAD, 79.6\%) replicate outside BRCA. CCND1 (STAD, 57.1\%) and MDM2 (STAD, 42.0\%) do not, unlike in BRCA where both replicate; ERBB2 shows the same cancer-type-dependent pattern in the opposite direction -- anti-concordant in BRCA and COAD (18.0\%, 14.3\%) but concordant in STAD (93.6\%). CCND2 (COAD, 54.2\%) also fails to replicate, but unlike CCND1 and MDM2 it never validated in BRCA either -- the only proliferation-machinery gene in the panel that fails in every cancer type tested. Section~\ref{sec:discussion} discusses this split as a hedged, post-hoc hypothesis, not an established mechanism.

The concordance result above validates CASCADE's predictions once correctly invoked; we separately test the step upstream of that, whether an LLM-based agent correctly grounds a natural-language request into CASCADE's real MCP tool-call parameters. Across 35 hand-labeled queries spanning gene aliases, informal cancer-type names, non-canonical perturbation phrasing, implicit/default parameters, and multi-entity distractors, CASCADE's own documented local-deployment default (\texttt{llama3.1:8b} via Ollama) reaches 71.4\% exact parameter-match on its raw output, and a larger local model (\texttt{qwen2.5:72b}) reaches 85.7\%. The gap is not uniform: schema-adherence failures (a required parameter left unset or misrouted) occur almost entirely with the smaller model and are fully resolved by the larger one, and gene-alias failures -- though model-size-invariant in each model's raw output -- resolve correctly for both models once CASCADE's own \texttt{resolve\_alias()} step is applied server-side (the same step CASCADE's real MCP server runs before any network lookup). A third failure mode does not resolve, however: both models confidently default to the wrong perturbation type on maximally ambiguous queries, and a targeted server-side fix cannot help, because its trigger condition -- the field being left unset -- never actually occurs (Section~\ref{sec:agent_grounding_results}).
\end{abstract}

\section{Introduction}

Predicting the downstream consequences of perturbing a gene is a central task in cancer systems biology, informing both mechanistic hypotheses and therapeutic target prioritization \citep{califano2017}. Tools built on precomputed regulatory networks can generate such predictions cheaply, without new experimental data or model training -- but this efficiency comes at a validation cost. The strongest evidence would be a held-out experimental perturbation screen; lacking that, network-propagation tools are typically validated instead against cancer-gene reference sets, the approach RegNetAgents \citep{regnetagents} itself takes, via enrichment for known driver genes among candidate genes. That check only asks whether a predicted candidate happens to already be a known cancer gene (\emph{membership} validation) -- it does not ask whether the prediction's specific directional claim, this gene goes up or that one goes down, matches what actually happens in real disease. This distinction has real consequences: when such a tool is exposed as a callable function to an LLM-based agent, as CASCADE is via MCP, the agent's report to a user inherits whatever validation gap the underlying tool carries, and membership validation cannot close it, since it never checks the specific directional claim the agent goes on to repeat.

RegNetAgents \citep{regnetagents} tested whether a focal gene's upstream regulator candidates, classified by network origin, are enriched for known cancer driver genes; against random genes, tumor-network candidates showed a modest excess (consistent in COAD, borderline in BRCA), and population-averaged (GREmLN-only) candidates showed none. That validation strategy is well suited to a regulator-\emph{identification} task, where the object of interest is a ranked list of candidate genes to be cross-referenced against a driver-gene list.

CASCADE addresses a different, complementary task: given a focal gene and a perturbation type (knockdown or overexpression), predict the \emph{downstream} transcriptional consequences by propagating a signed perturbation signal through a precomputed regulatory network. This is architecturally similar to RegNetAgents -- both are implemented as LangGraph-orchestrated workflows over precomputed ARACNe networks, exposed via MCP for natural-language queries -- but the object being validated is fundamentally different: not "is this candidate a known cancer gene" but "does this specific predicted direction of change actually happen." Driver-gene enrichment, the validation strategy used for RegNetAgents, is a weaker test for this kind of claim, because it can only check whether predicted genes are plausible cancer genes in general, not whether the predicted \emph{direction} of change is correct. We instead validate against real patient tumor data directly, so that the specific claim an agent surfaces when it calls this tool is one we have directly checked against reality, not merely inferred to be plausible from network topology or gene-set membership.

CASCADE's downstream perturbation-prediction task relates to a broader literature on computational perturbation-effect prediction. VIPER \citep{viper}, built on the same ARACNe\slash regulatory-network lineage as CASCADE's networks, infers protein activity from regulon enrichment in observed gene expression -- the reverse of CASCADE's task, which predicts the transcriptional consequence of a hypothetical perturbation rather than inferring activity from data already observed. GEARS \citep{gears} and CPA \citep{cpa} address a task closer to CASCADE's: predicting transcriptional responses to gene perturbations, including unseen combinations. Both are learned models trained on large single-cell perturbational screens, giving them access to combinatorial and dose-dependent effects that CASCADE's static network propagation does not model, at the cost of requiring substantial training data and infrastructure that CASCADE's precomputed-network approach avoids. We are not aware of a prior application of any of these approaches to real patient tumor genotype-expression data in the manner reported here; this paper's contribution is a validation methodology for CASCADE specifically, not a claim that its propagation-based predictions are more accurate than learned alternatives -- a head-to-head comparison we did not attempt and consider open future work.

\subsection{Contribution}

This paper makes two related but distinct contributions: a patient-data concordance methodology for validating CASCADE's downstream perturbation predictions, and a benchmark of whether an LLM-based agent correctly grounds natural-language requests into CASCADE's real MCP tool-call parameters:

\begin{enumerate}
\item A directional concordance methodology -- focal-gene copy-number amplification as a proxy for the inverse of knockdown, tested against real patient genotype-expression data with permutation-based significance -- that validates downstream perturbation predictions independent of curated cancer-gene annotation, and generalizes in principle to any regulatory-network-based perturbation tool.
\item For MYC: a strong, permutation-controlled concordance signal that survives a PAM50 subtype control and replicates in an independent cohort (METABRIC) with no data-provenance overlap with CASCADE's networks.
\item A fifteen-gene generalization test showing this validation is gene-specific: nine proliferation-machinery regulators replicate (seven cleanly, two weakly) while six do not -- four lineage-identity transcription factors, a receptor tyrosine kinase, and CCND2, a cyclin-D paralog of two validating genes. Section~\ref{sec:discussion} discusses the resulting split as a post-hoc hypothesis, not an established mechanism.
\item A benchmark of whether an LLM-based agent correctly grounds natural-language requests into CASCADE's real MCP tool-call parameters, distinct from whether CASCADE's predictions are accurate once invoked: schema-adherence failures are model-size-dependent, gene-alias failures are resolved server-side by CASCADE's alias table, and a third failure mode -- confidently defaulting to the wrong perturbation type on maximally ambiguous queries -- persisted through a targeted fix, because both models tested always supply a guess rather than leaving the field unset (Section~\ref{sec:agent_grounding_results}).
\end{enumerate}

Section~\ref{sec:agent_grounding} (methodology) and Section~\ref{sec:agent_grounding_results} (results) cover the agentic tool-call grounding contribution; the remaining Methods and Results subsections cover the patient-data validation methodology and results -- ARACNe network construction, TCGA/METABRIC cohort details, and gene-panel selection -- and assume more domain background.

\section{Methods}

\subsection{CASCADE architecture}
\label{sec:architecture}

CASCADE is a Python package exposing gene perturbation analysis through a Lang\allowbreak Graph-orchestrated workflow and an MCP server, following the same general orchestration pattern as RegNetAgents \citep{regnetagents, langgraph}: a directed workflow classifies a focal gene's regulatory role from precomputed network topology, routes to appropriate analyses, and executes independent sub-analyses concurrently before assembling a structured report. Every result in this paper uses CASCADE's actual default perturbation-propagation behavior for TCGA networks: network propagation blended with GREmLN embedding-based cosine similarity \citep{gremln} via a weighted parameter ($\alpha=0.7$), with automatic fallback to network-only propagation only if the embedding model is unavailable (not the case for any result reported here). Every experiment in this paper invokes \texttt{Cascade\allowbreak Workflow\allowbreak.run()} (below) directly. The network component is a deterministic breadth-first algorithm over the ARACNe network's directed regulator$\to$target edges: for a knockdown, the focal gene receives an initial effect of $-1.0$, and at each hop the effect propagated to a target gene is the parent gene's current effect multiplied by the edge weight and a decay factor of $0.5$, accumulating additively across paths. For TCGA networks, the edge weight is the ARACNe mutual-information magnitude signed by the edge's mode-of-action annotation (activating edges preserve sign, repressive edges flip it) -- this sign-flipping is the primary mechanism producing the up/down direction predictions the concordance test in Section~\ref{sec:methods_concordance} evaluates.

The embedding component adds cosine-similarity-weighted contributions from CASCADE's pre-trained gene embeddings on top of this network signal, and can surface additional candidate genes reachable via embedding similarity but not network propagation; because $\alpha=0.7$ keeps the network-derived signal dominant, embedding blending rescales magnitude and can add new candidates but does not flip the direction of genes already reachable via the network. All analyses in this paper use CASCADE's own default propagation depth (2 hops) and default result size (top 25 by $|\text{effect}|$ for the internal role-classification pathway; the validation experiments described below instead use the top 50 by $|\text{effect}|$, as specified per experiment).

CASCADE supports ARACNe regulatory networks for 14 TCGA cancer types from the Bioconductor \texttt{aracne.networks} package \citep{aracnenetworks}, built from TCGA RNA-seq data downloaded April 2015 using ARACNe-AP \citep{lachmann2016}; users install them locally from the package authors' Zenodo record (CC BY-NC-ND 4.0) \citep{aracnezenodo} with an installer script. Networks are symbol-native (no Ensembl mapping required for TCGA network operations) and carry per-edge mode-of-action annotation (activating/repressive) derived from the same source.

Before propagation, CASCADE's workflow classifies the focal gene's regulatory role from network topology (Table~\ref{tab:role_rules}), which determines downstream routing within the full CASCADE system (e.g.\ whether regulator analysis or protein-interaction evidence is prioritized). This classification is deterministic and threshold-based, not learned; it is reported here for completeness of the architecture description, not as a claim evaluated in this paper. Because this routing logic is deterministic code rather than a learned or LLM-driven decision, we verify its correctness via conventional unit tests (\texttt{tests/test\_workflow.py}, covering role classification and batch-dispatch routing across gene-role and depth combinations) rather than an empirical benchmark; the agent tool-call grounding benchmark (Section~\ref{sec:agent_grounding}) is reserved for the one step in CASCADE's pipeline that does involve genuine model judgment under ambiguity.

\begin{table}[H]
\centering
\small
\caption{CASCADE gene-role classification rules (\texttt{cascade\_langgraph\_workflow.py}).}
\label{tab:role_rules}
\begin{tabular}{ll}
\toprule
Gene role & Rule (network out-/in-degree) \\
\midrule
Master regulator & Downstream targets $>50$ \\
Transcription factor & $10 <$ downstream targets $\leq 50$ \\
Minor regulator & $0 <$ downstream targets $\leq 10$ \\
Effector & 0 downstream targets, $\geq 1$ upstream regulator \\
Isolated & 0 downstream targets, 0 upstream regulators \\
\bottomrule
\end{tabular}
\end{table}

\begin{figure}[H]
\centering
\includegraphics[width=0.7\textwidth]{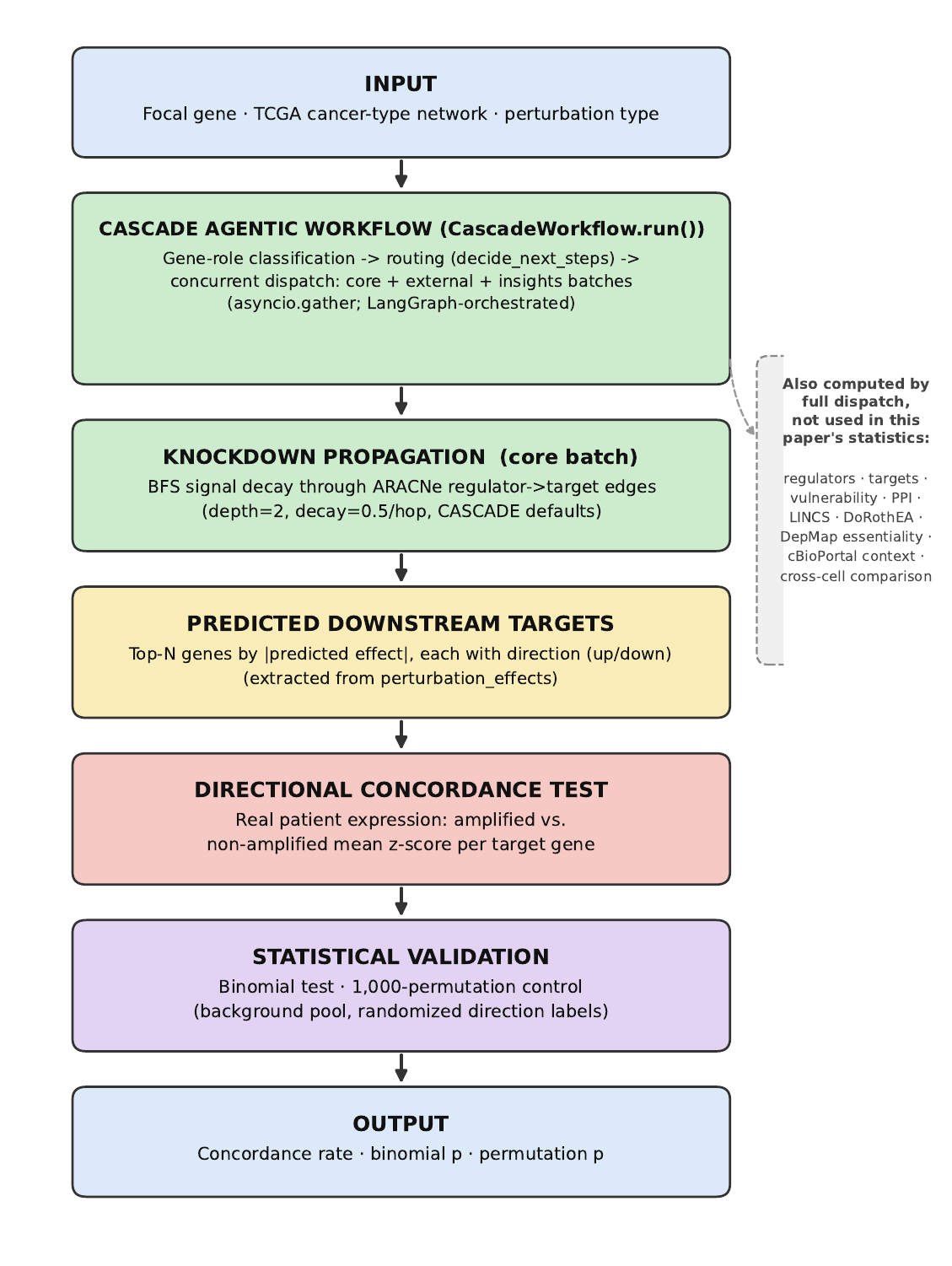}
\caption{Pipeline exercised by this paper's experiments: focal-gene knockdown propagation via CASCADE's agentic entry point, \texttt{Cascade\allowbreak Workflow\allowbreak.run()}, producing direction-labeled predicted targets, tested for directional concordance against real patient copy-number/expression data, with binomial and permutation-based statistical validation. CASCADE's own routing logic concurrently computes additional non-LLM analyses as a byproduct of this call (grey panel); these are not used in this paper's statistics.}
\label{fig:pipeline}
\end{figure}

Figure~\ref{fig:pipeline} summarizes this pipeline end to end. \texttt{\_decide\_next\_steps} computes a role- and depth-conditional required-analysis set and routes the workflow's concurrent batch dispatch to cover exactly that set (detailed below); an optional LLM-based node can synthesize a narrative interpretation of a completed report without altering any of its underlying scores, but this LLM node (\texttt{include\_llm\_insights}) is not enabled in any experiment in this paper.

CASCADE's public agentic API, and the exact call used directly by every experiment script in this paper, is:

\begin{verbatim}
from cascade_langgraph_workflow import CascadeWorkflow

workflow = CascadeWorkflow()
report = await workflow.run(
    gene="MYC",
    perturbation_type="knockdown",
    analysis_depth="focused",
    network_source="tcga",
    tcga_network="brca",
    top_k=50,
)
predicted_targets = report["perturbation_effects"]["top_affected_genes"]
\end{verbatim}

\texttt{top\_k} is discussed further below; every argument shown here is part of CASCADE's public API.

We verified empirically, not merely by reading the routing code, exactly which of CASCADE's internal analyses \texttt{analysis\_depth="focused"} triggers for the gene roles tested in this paper. \texttt{\_decide\_\allowbreak next\_\allowbreak steps} classifies the focal gene's regulatory role and, for \texttt{"focused"}, requires \texttt{perturbation} plus either \texttt{targets} (master-regulator and transcription-factor genes) or \texttt{ppi} (other roles); every gene-cancer-type combination tested in this paper is classified as master regulator or transcription factor (Table~\ref{tab:role_applied}), so every call here requires exactly \{\texttt{perturbation}, \texttt{targets}\}, both members of the same core-analysis batch group, and routing sends the call to that single batch node rather than the full three-way concurrent dispatch. Only the \texttt{perturbation\_effects} field of the returned report is extracted and used in this paper's statistical tests; the concurrently-computed \texttt{targets} analysis is not evaluated further and is not reported as a result in this paper.

\texttt{Cascade\allowbreak Workflow\allowbreak.run()} accepts a \texttt{top\_k} parameter (default 25, CASCADE's single-query default), threaded through to every internal propagation call site; this paper's $N=50$ concordance panels pass \texttt{top\_k=50} explicitly. It likewise accepts a \texttt{propagation\_depth} parameter (default 2, CASCADE's default BFS hop count), threaded through the same call sites; every result in this paper uses this default depth of 2. Every experiment in this paper -- the MYC primary panels, the PAM50 and METABRIC checks, and the fifteen-gene generalization panel (Section~\ref{sec:gene_specificity}) -- uses \texttt{analysis\_depth="focused"}; none uses \texttt{"comprehensive"}, so the additional analyses \texttt{"comprehensive"} would require for master-regulator\slash transcription-factor genes (\texttt{vulnerability}, \texttt{lincs}, \texttt{regulators}, \texttt{similar}, \texttt{dorothea}, \texttt{depmap}, \texttt{cbioportal}) are not computed by any result reported here. Table~\ref{tab:role_applied}'s role classifications are also examined descriptively in Results (Section~\ref{sec:gene_specificity}) against this paper's biological concordance findings, beyond their role in determining routing here -- though not as a systematically tested variable, as that section itself notes.

\FloatBarrier
\subsection{Directional concordance methodology}
\label{sec:methods_concordance}

For a focal gene $g$ in a given TCGA cancer-type network, CASCADE's knockdown propagation yields a signed effect for every reachable downstream gene. We take the top $N=50$ genes by $|\text{effect}|$ and record each gene's predicted direction (down if effect $<0$, up if effect $>0$).

We treat focal-gene copy-number amplification (GISTIC discrete value $=2$) as an approximate dosage-based proxy for increased focal-gene activity -- the inverse, in direction, of a knockdown. For each predicted target gene, we compare its mean mRNA expression $z$-score between amplified and non-amplified (GISTIC $=0$) patient samples. A gene predicted to go \emph{down} upon knockdown (i.e.\ positively regulated by the focal gene) is scored \emph{concordant} if its mean expression is \emph{higher} in amplified samples; a gene predicted to go \emph{up} (negatively regulated) is concordant if its mean expression is \emph{lower} in amplified samples. Genes with fewer than 10 samples in either group are excluded.

Significance is assessed with a one-tailed binomial test (null: concordance rate $=0.5$) and an independent permutation control: 1,000 draws of $N$ genes from a background pool of 200 randomly sampled network genes, excluding the focal gene itself and every gene in CASCADE's own predicted-target set for that run (so the null distribution cannot be contaminated by the real signal it is meant to be compared against), with real amplified/non-amplified expression differences precomputed, each draw randomly assigned a predicted-direction label matching the true up/down proportion of the real predicted set. The empirical $p$-value is the fraction of permuted concordance rates meeting or exceeding the observed rate. This directly tests whether the observed concordance rate exceeds what arbitrary genes, given the same class balance, would produce by chance.

\subsection{Focal gene panel selection for generalization testing}
\label{sec:panel_selection}

Beyond MYC, we sought additional focal genes to test whether the directional concordance methodology generalizes. Because the methodology requires focal-gene amplification as the dosage proxy, candidate genes were required to satisfy two independent criteria in a given TCGA network: (i) network out-degree $\geq 25$ (so the predicted-target panel reaches full size), and (ii) at least $\sim$15--20 GISTIC-amplified patient samples in the corresponding cBioPortal PanCancer Atlas study (so amplified/non-amplified group means are estimable). We screened MYC, ERBB2, and seven additional transcription factors with documented amplification-driven oncogenic roles in at least one tissue (CTNNB1, GATA3, FOXA1, KLF5, SOX9, E2F3, MYB) across BRCA, COAD, and STAD.

Most lineage-restricted transcription factors failed criterion (ii) in at least two of the three cancer types -- e.g.\ CTNNB1 is activated predominantly by point mutation rather than amplification and had at most one amplified sample in any tested cancer type; GATA3 and FOXA1 similarly lacked sufficient amplified samples outside BRCA. Only MYC and ERBB2 satisfied both criteria in all three cancer types; E2F3, SOX9, GATA3, and FOXA1 satisfied both criteria in BRCA only. This selection process is itself a finding: strong, cross-tissue, amplification-driven dosage effects comparable to MYC are genuinely rare among transcription factors, which constrains the achievable size of any generalization panel built on this methodology.

\subsection{Data sources}

Real-patient expression and copy-number data are drawn from the TCGA PanCancer Atlas 2018 \citep{tcga_pancan}, with copy-number calls generated by GISTIC2.0 \citep{mermel2011}, accessed via the cBioPortal REST API \citep{cerami2012, gao2013}. This API's data is GDC-harmonized against GRCh38 -- a different reference genome and processing pipeline than the 2015 raw-download data underlying CASCADE's ARACNe networks, though very likely drawing on an overlapping TCGA patient cohort. PAM50 molecular subtype annotations were obtained from the same cBioPortal study's patient-level clinical data. Independent-cohort replication uses METABRIC \citep{curtis2012, pereira2016}, a breast cancer cohort of 2,509 tumors profiled by microarray at UK and Canadian institutions, with no data-provenance relationship to TCGA.

All fetches use cBioPortal's batch endpoints (\texttt{genes/fetch} for symbol$\to$Entrez resolution, \texttt{molecular-data/fetch} for multi-gene expression retrieval), making each experiment a small, fixed number of API calls regardless of gene panel size. Analyses are otherwise read-only scripts against local network files and public APIs.

\subsection{Agent tool-call grounding methodology}
\label{sec:agent_grounding}

Every experiment above calls \texttt{CascadeWorkflow.run()} directly with hand-specified parameters, which validates CASCADE's predictions but not the step upstream of them: whether an LLM-based agent, given only a natural-language request and CASCADE's real MCP tool schema, correctly fills in the parameters that call actually needs.

This complements, at much smaller scale, an existing benchmark literature on LLM tool-use and function-calling generally: Gorilla \citep{gorilla} and the Berkeley Function-Calling Leaderboard \citep{bfcl} evaluate schema-adherence and API-selection accuracy across large synthetic or aggregated API collections. The benchmark below instead tests end-to-end natural-language grounding against one real, deployed tool's actual MCP schema, on 35 queries constructed specifically to probe CASCADE's own failure surface (gene aliases, informal cancer-type names, ambiguous perturbation direction) rather than to sample broadly across unrelated APIs.

We test this separately and explicitly, using CASCADE's actual \texttt{comprehensive\_\allowbreak perturbation\_\allowbreak analysis} tool definition (verbatim from \texttt{cascade\_langgraph\_mcp\_server.py}) against 35 hand-labeled natural-language queries spanning seven categories (five queries each): baseline TCGA-cancer-type requests, baseline immune-cell-type requests, gene aliases (a deliberate mix of aliases resolvable via CASCADE's alias table and one resolvable only via general reasoning, to test both cases, Section~\ref{sec:agent_grounding_results}), informal/full cancer-type names, non-canonical perturbation-verb phrasing (``silence,'' ``boost,'' ``suppress,'' ``delete function''), queries that omit parameters entirely (testing whether the agent falls back to CASCADE's documented defaults), and queries containing a distractor gene or cancer type not actually requested. Ground truth for each query is the parameter set CASCADE's tool would need to receive to behave correctly, with an omitted field scored against the default CASCADE's own schema documents for it (e.g.\ \texttt{perturbation\_type} defaults to \texttt{knockdown}), not against a bare absence.

Two models are tested via a local Ollama server: \texttt{llama3.1:8b}, CASCADE's own documented default for its separate LLM-insights feature (Section~\ref{sec:architecture}) and the more representative test of what a fully local/offline CASCADE deployment would use, and \texttt{qwen2.5:72b-instruct-q4\_0}, a substantially larger local model, reported as a secondary comparison rather than the primary result to avoid presenting only the model that performs best. Each query is sent once per model with the real tool schema attached; the model's emitted tool-call arguments are compared field-by-field against ground truth.

CASCADE resolves a small, fixed set of common informal gene names server-side (e.g.\ \texttt{HER2}$\to$\texttt{ERBB2}) via \texttt{tools/gene\_id\_mapper.py}'s \texttt{resolve\_alias()} function, before any network lookup. Because a model can still emit an alias's literal informal name even though CASCADE's table can resolve it server-side, we score each gene-alias query two ways: an exact match against the model's raw tool-call output, and a second match after running the model's \texttt{gene} value through CASCADE's real \texttt{resolve\_alias()} function, the same step its MCP server runs before any network lookup (Section~\ref{sec:agent_grounding_results}).

We also implemented a targeted fix for the perturbation-type default-handling failure this benchmark surfaced (Section~\ref{sec:agent_grounding_results}): an optional \texttt{query} parameter added to \texttt{comprehensive\_\allowbreak perturbation\_\allowbreak analysis}'s schema, carrying the caller's original natural-language request, and a check in the server's handler that scans this text for directional cues (e.g.\ ``knock down''/``silence'' vs.\ ``overexpress''/``boost'', matched via a small regex list) whenever \texttt{perturbation\_type} is omitted from the tool call, returning a \texttt{clarification\_needed} response instead of applying the \texttt{knockdown} default when no single direction is detected; an explicitly-supplied \texttt{perturbation\_type} value is never overridden. This detection is a simple keyword scan and has a known blind spot we did not exercise in this benchmark: it cannot attribute a directional cue to a specific gene, so a multi-entity query naming two genes with different directions (e.g.\ ``knocked down GATA3, now overexpress FOXA1'') would read as conflicting cues even though the request is unambiguous once each verb is attributed to its gene. None of the 35 queries tested exercises this case, so we report it here as an acknowledged gap in the detection logic rather than as a tested limitation of the results below.

\section{Results}

\subsection{MYC knockdown predictions show strong, multi-cancer-type concordance with real patient data}
\label{sec:myc_main}

Table~\ref{tab:myc_main} summarizes the primary result, using CASCADE's actual default embedding-enhanced propagation ($\alpha=0.7$; Section~\ref{sec:architecture}). Across all three tested TCGA cancer types, CASCADE's top-50 predicted MYC-knockdown targets show concordance rates 24--36 percentage points above their respective permutation baselines, each with permutation-empirical $p=0.0000$ (no permutation among 1,000 draws matched or exceeded the observed rate). Zero candidates in any cancer type were embedding-only additions (genes reachable via embedding similarity but not network propagation), consistent with MYC's broad network connectivity already spanning the genes embedding similarity would otherwise surface.

\begin{table}[h]
\centering
\caption{MYC knockdown concordance with real patient MYC-amplification status, by TCGA cancer type.}
\label{tab:myc_main}
\begin{tabular}{lccccc}
\toprule
Cancer type & Concordant & Rate & Binomial $p$ & Permutation mean & Permutation $p$ \\
\midrule
BRCA & 45/50 & 90.0\% & $<10^{-8}$ & 53.8\% & 0.0000 \\
COAD & 36/50 & 72.0\% & 0.0013 & 48.4\% & 0.0000 \\
STAD & 42/49 & 85.7\% & $<0.0001$ & 50.4\% & 0.0000 \\
\bottomrule
\end{tabular}
\end{table}

The concordant gene set in BRCA is biologically coherent, not merely statistically significant: top concordant genes include NOP14, NOB1, RIOK1, NOLC1, WDR43, TAF1D, SNRPD1, and KAT2A -- ribosome biogenesis and nucleolar machinery components, MYC's best-established transcriptional program in cancer biology \citep{vanriggelen2010}. This independent biological plausibility check is consistent with the statistical result reflecting genuine regulatory signal rather than an artifact of the test design.

The result is not sensitive to the arbitrary choice of panel size $N=50$: repeating the identical BRCA/MYC test at $N=25$ and $N=100$ yields 22/25 (88.0\%, binomial $p=0.0001$) and 92/100 (92.0\%, binomial $p<0.0001$) concordant respectively, both with permutation-empirical $p\leq0.001$ -- concordance rates of 88.0\%, 90.0\%, and 92.0\% at $N=25$, $50$, and $100$ respectively, essentially flat across a fourfold range in panel size.

Table~\ref{tab:myc_worked} shows the fifteen highest-magnitude predicted targets from a single MYC/BRCA query, illustrating the practical output the concordance test evaluates: for each predicted target, its CASCADE-predicted direction and its actual mean expression $z$-score in MYC-amplified versus non-amplified patients. This top-15 slice's 13/15 (86.7\%) ``down'' share is not representative of the full top-50 panel, which is considerably more balanced (Appendix~\ref{app:baseline}).

\begin{table}[t]
\centering
\small
\caption{Top 15 CASCADE-predicted MYC knockdown targets (BRCA), by $|\text{effect}|$, with real patient expression comparison.}
\label{tab:myc_worked}
\begin{tabular}{lcccc}
\toprule
Gene & Predicted direction & Amplified mean $z$ & Non-amplified mean $z$ & Concordant \\
\midrule
TWNK & down & $+0.515$ & $-0.419$ & Yes \\
PDCD11 & down & $+0.405$ & $-0.361$ & Yes \\
NOP14 & down & $+0.248$ & $-0.389$ & Yes \\
NOB1 & down & $-0.040$ & $-0.654$ & Yes \\
SCP2 & up & $-0.234$ & $+0.091$ & Yes \\
GEMIN4 & down & $-0.225$ & $-0.606$ & Yes \\
RIOK1 & down & $+1.246$ & $-0.253$ & Yes \\
SNHG29 & down & $-0.199$ & $-0.164$ & \textbf{No} \\
TAF1D & down & $+0.150$ & $-0.390$ & Yes \\
SNRPD1 & down & $+0.719$ & $-0.400$ & Yes \\
NOLC1 & down & $+0.592$ & $-0.502$ & Yes \\
WDR43 & down & $+0.980$ & $-0.282$ & Yes \\
COQ8A & down & $+1.244$ & $+0.403$ & Yes \\
TRIM62 & up & $-0.375$ & $-0.013$ & Yes \\
KAT2A & down & $+0.264$ & $-0.198$ & Yes \\
\bottomrule
\end{tabular}
\vspace{2pt}

\begin{minipage}{\textwidth}
\footnotesize Illustrative top-15-by-magnitude slice; its 86.7\% ``down'' share is not representative of the full 50-gene panel's up/down balance, which is considerably more balanced (Appendix~\ref{app:baseline}).
\end{minipage}
\end{table}

\begin{figure}[h]
\centering
\includegraphics[width=0.95\textwidth]{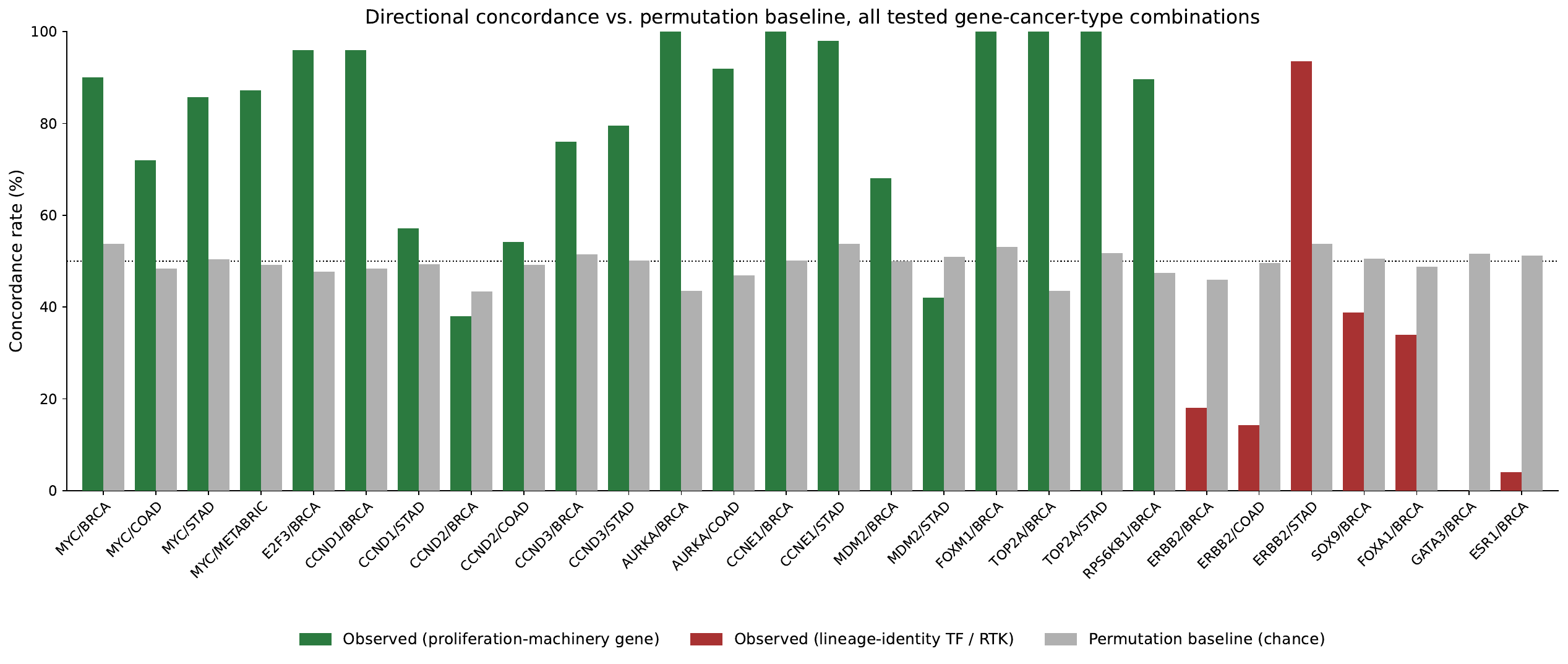}
\caption{Observed concordance rate (colored bars) versus permutation-derived chance baseline (grey bars), for every gene-cancer-type combination tested, including the cross-cancer-type extension for AURKA, CCND1, CCND2, CCND3, CCNE1, MDM2, and TOP2A. Green bars: proliferation-machinery-associated genes (MYC, E2F3, CCND1, CCND2, CCND3, AURKA, CCNE1, MDM2, FOXM1, TOP2A, RPS6KB1). Red bars: lineage-identity transcription factors (SOX9, FOXA1, GATA3, ESR1) or the receptor tyrosine kinase ERBB2. Bar color reflects gene category, not whether that individual result was concordant -- e.g.\ ERBB2/STAD is colored red (category) despite exceeding its own permutation baseline (result), and conversely CCND1/STAD, CCND2/BRCA, CCND2/COAD, and MDM2/STAD are colored green (category) despite not being significantly concordant (result); all instances of this inconsistency are discussed in Section~\ref{sec:gene_specificity}. MYC is concordant in all four contexts tested (three TCGA cancer types plus the independent METABRIC cohort); the proliferation-machinery group validates in most but not all gene-cancer-type combinations (AURKA, CCNE1, TOP2A, and CCND3 in a second cancer type; CCND1 and MDM2 only in BRCA; CCND2 not at all, in either cancer type tested, despite its paralogs CCND1 and CCND3 both validating) while the lineage-identity/RTK group shows mixed to strongly anti-concordant results. ESR1 was selected after, and specifically to test, the proliferation-machinery/lineage-identity hypothesis discussed in Section~\ref{sec:discussion} -- the only gene in this figure chosen because it was predicted to fail rather than to validate.}
\label{fig:summary}
\end{figure}

Figure~\ref{fig:summary} summarizes observed concordance rates against permutation baselines for every gene-cancer-type combination tested in this paper, including the additional-gene generalization tests reported in Section~\ref{sec:gene_specificity}.

\FloatBarrier
\subsection{A baseline comparison against curated, identity-matched gene sets}
\label{sec:baseline_comparison}

A positive concordance result does not by itself show that CASCADE's specific algorithm adds value beyond a public, non-CASCADE gene list, so we checked this directly against two curated, identity-matched gene lists -- genes with well-established roles as transcriptional activators (MYC, and the activating E2Fs including E2F3), such that predicting their target-set members uniformly ``down'' upon knockdown is a biologically grounded assumption rather than an artificial convention invented for comparability -- using each list's full resolvable gene set and comparing its concordance rate against CASCADE's own via Fisher's exact test, which accounts for the two groups' different sample sizes directly rather than treating both rates as equally precise (full methodology and results in Appendix~\ref{app:baseline}, Table~\ref{tab:baseline_comparison}).

Against an exhaustive curated MYC-target gene list, the gap is narrow and inconsistent in direction -- 1--4 percentage points across all three cancer types, with the curated list ahead in BRCA and STAD and CASCADE ahead in COAD -- and not statistically significant in any of the three (Fisher's exact $p=0.34$--$0.86$): CASCADE's overall accuracy is not established to exceed an independent expert-curated database, but equally not established to fall short of it. For E2F3, CASCADE's rate is nominally higher than its curated comparison list (96.0\% vs.\ 90.7\%), but this difference is not statistically significant either ($p=0.384$). None of the four comparisons survives Benjamini-Hochberg correction across all four ($q=0.767$--$0.863$, Table~\ref{tab:baseline_comparison}). This correction is applied within its own four-test family, separate from the twenty-seven-combination correction in Table~\ref{tab:other_genes}: the two test different claims -- baseline superiority over a curated gene list here, versus cross-gene generalization there -- so we do not pool them into a single correction family.

This is a separate question from whether CASCADE's own direction-calling matters: forcing every predicted gene to ``down'' lowers concordance by 14--29 percentage points in every panel tested, so CASCADE's gene-specific reasoning is doing substantial, measurable work relative to a naive version of itself. Taken together, this comparison shows CASCADE's MYC and E2F3 predictions are correct and non-trivially derived, not that CASCADE's overall accuracy is established to exceed existing public knowledge of MYC- or E2F-driven biology in either of the two gene-identity comparisons tested.

\FloatBarrier
\subsection{The result survives a PAM50 subtype control}
\label{sec:pam50}

MYC amplification is not uniformly distributed across PAM50 molecular subtypes in our BRCA cohort (Basal 61 amplified/13 non-amplified; LumA 36/251; LumB 32/33; Her2 18/20; Normal 6/21), raising the possibility that predicted-target concordance reflects subtype composition differences rather than MYC dosage itself. Restricting the identical test to BRCA\_LumB alone -- the subtype with the most balanced amplified/non-amplified split, and therefore the best-powered within-subtype test available -- yielded 46/50 concordant (92.0\%, binomial $p=2.2\times10^{-10}$), a result that \emph{strengthened} rather than weakened relative to the full unstratified cohort. This is inconsistent with a pure subtype-confound explanation.

\subsection{Independent-cohort replication rules out a shared-data artifact}
\label{sec:metabric}

CASCADE's TCGA ARACNe networks and the TCGA PanCancer Atlas expression/copy-number data used above are both ultimately derived from the TCGA-BRCA cohort, processed through different pipelines at different times (2015 raw download vs.\ 2018 GDC-harmonized GRCh38) but very likely overlapping in the underlying patients. To rule out the possibility that real patient-specific biology shared between these two data releases was inflating apparent validation, we repeated the identical concordance test against METABRIC \citep{curtis2012, pereira2016} -- a cohort with no data-provenance relationship to TCGA whatsoever (different patients, different countries, microarray rather than RNA-seq quantification). CASCADE's TCGA-network-derived MYC predictions were tested, unmodified, against METABRIC's amplified ($n=554$) and non-amplified ($n=1{,}110$) patient expression data: 41/47 testable genes were concordant (87.2\%, binomial $p<10^{-6}$, permutation-empirical $p=0.0000$, permutation mean 49.2\%) -- a result essentially identical in magnitude to the original TCGA-BRCA finding (90.0\%).

\FloatBarrier
\subsection{Validation is gene-specific, not universal}
\label{sec:gene_specificity}

We tested fifteen additional genes for generalization (Table~\ref{tab:other_genes}). Genes entered the panel three ways, disclosed here against any appearance of post-hoc cherry-picking: an initial blind screen against the eligibility criteria in Section~\ref{sec:panel_selection} (E2F3, ERBB2, SOX9, FOXA1, GATA3); systematic screens of proliferation-machinery/cell-cycle candidates against those same unchanged criteria, testing every eligible gene regardless of expected outcome (CCND1, AURKA, CCNE1, MDM2, CCND3, CCND2, FOXM1, TOP2A, RPS6KB1; four further candidates screened -- CDK4, MYBL2, E2F1, BIRC5 -- failed eligibility and were not tested); and one gene chosen specifically to falsify rather than confirm the working hypothesis, ESR1, selected after the proliferation-machinery/lineage-identity hypothesis (Section~\ref{sec:discussion}) had already formed, because that hypothesis predicted it would fail (it did, Table~\ref{tab:other_genes}).

Of the ten proliferation-machinery genes tested, nine validate somewhere (E2F3, CCND1, CCND3, AURKA, CCNE1, MDM2, FOXM1, TOP2A, RPS6KB1; 68--100\%, Table~\ref{tab:other_genes}) and only CCND2 does not. None of the five lineage-identity/RTK genes validate as a category (GATA3, FOXA1, SOX9, ESR1, ERBB2) -- though ERBB2's STAD result is a significant exception within that group (93.6\%, $q<0.0001$; discussed below). Transcription-factor status does not separate the two groups -- Section~\ref{sec:discussion} walks through the gene-by-gene split, including how it differs from CASCADE's own network-derived role classification for some of these genes.

\textbf{CCND2 fails outright, unlike its paralogs, in both cancer types tested.} CCND2 is the sharpest exception to any clean gene-class rule: eligible under criteria identical to CCND1 and CCND3 (16 amplified samples, network out-degree 34), it fails outright (19/50, 38.0\%, $p=0.9675$, below its own 43.4\% permutation baseline) in BRCA, the cancer type where both paralogs validate. Extended to COAD -- where its out-degree (48) and amplified-sample count (16) independently clear the same eligibility criteria -- CCND2 again fails to validate (26/48, 54.2\%, $p=0.333$, essentially chance), while CCND3 replicates cleanly in its own second cancer type, STAD (39/49, 79.6\%, $p<0.0001$). CCND2 is therefore the only proliferation-machinery gene in the panel that fails to validate in every cancer type tested.

\textbf{The cross-cancer-type extension surfaces further inconsistency.} Seven non-MYC/ERBB2 proliferation-machinery genes were extended to a second cancer type wherever eligibility criteria were met (Section~\ref{sec:panel_selection}): AURKA (COAD); CCND1, CCNE1, MDM2, and TOP2A (STAD); and CCND2 and CCND3 (COAD and STAD respectively, above). Four replicate outside BRCA (AURKA 92.0\%, CCNE1 98.0\%, TOP2A 100.0\%, each $p<0.0001$; CCND3 79.6\%, $p<0.0001$), but three do not: CCND1 falls from 96.0\% in BRCA to 57.1\% in STAD ($p=0.196$), MDM2 falls from 68.0\% in BRCA to 42.0\% in STAD ($p=0.899$, below its own permutation baseline), and CCND2 fails in both cancer types tested (above). ERBB2 shows the same pattern even more starkly across all three cancer types tested: strong anti-concordance in BRCA and COAD (18.0\%, 14.3\%) but strong concordance in STAD (93.6\%). Gene class alone is therefore not sufficient to predict validation; cancer-type context matters too, for reasons this paper does not resolve.

\textbf{Ruling out an embedding-blend artifact.} Because CASCADE's default propagation blends network topology with GREmLN embedding similarity (Section~\ref{sec:architecture}), we checked whether that embedding component contributes unevenly across the two groups, which could confound the split reported above. It does not: embedding-only additions numbered 0 or 1 of 50 in every one of the twenty-seven gene-cancer-type combinations tested, with no difference between validating and failing groups. GATA3's result -- the most extreme in the panel (0/50 concordant) -- was checked further under network-only propagation, with the embedding component removed entirely: the result is an identical 0/50, confirming the anti-concordance is a property of the network and its annotations, not of CASCADE's embedding component.

\begin{table}[h]
\centering
\footnotesize
\caption{Concordance for additional focal genes tested for generalization, including cross-cancer-type extensions for five proliferation-machinery genes (Section~\ref{sec:gene_specificity}). Category is a hypothesis-driven grouping assigned by us (Section~\ref{sec:discussion}), not a CASCADE output and not a direct function of the Type column: genes of the same molecular Type can fall in different Categories (e.g.\ E2F3 and FOXM1 versus SOX9, FOXA1, GATA3, and ESR1 are all transcription factors by Type but split across both Categories).}
\label{tab:other_genes}
\setlength{\tabcolsep}{4pt}
\renewcommand{\arraystretch}{1.25}
\resizebox{\linewidth}{!}{%
\begin{tabular}{lllcccl}
\toprule
Gene & Category & Type & Cancer type & Concordance & BH-FDR $q$ & Assessment \\
\midrule
E2F3 & Proliferation-machinery & Transcription factor & BRCA & 96.0\% & $<0.0001$ & Replicates MYC \\
CCND1 & Proliferation-machinery & Cell-cycle regulator & BRCA & 96.0\% & $<0.0001$ & Replicates MYC \\
CCND1 & Proliferation-machinery & Cell-cycle regulator & STAD & 57.1\% & 0.2937 & Not significant \\
CCND2 & Proliferation-machinery & Cell-cycle regulator & BRCA & 38.0\% & 1.000 & Anti-concordant \\
CCND2 & Proliferation-machinery & Cell-cycle regulator & COAD & 54.2\% & 0.4728 & Not significant \\
CCND3 & Proliferation-machinery & Cell-cycle regulator & BRCA & 76.0\% & 0.0003 & Replicates (weaker) \\
CCND3 & Proliferation-machinery & Cell-cycle regulator & STAD & 79.6\% & $<0.0001$ & Replicates (weaker) \\
AURKA & Proliferation-machinery & Mitotic kinase & BRCA & 100.0\% & $<0.0001$ & Replicates MYC \\
AURKA & Proliferation-machinery & Mitotic kinase & COAD & 92.0\% & $<0.0001$ & Replicates MYC \\
CCNE1 & Proliferation-machinery & G1/S cyclin & BRCA & 100.0\% & $<0.0001$ & Replicates MYC \\
CCNE1 & Proliferation-machinery & G1/S cyclin & STAD & 98.0\% & $<0.0001$ & Replicates MYC \\
MDM2 & Proliferation-machinery & Ubiquitin ligase & BRCA & 68.0\% & 0.0122 & Replicates (weaker) \\
MDM2 & Proliferation-machinery & Ubiquitin ligase & STAD & 42.0\% & 1.000 & Not significant \\
FOXM1 & Proliferation-machinery & Transcription factor & BRCA & 100.0\% & $<0.0001$ & Replicates MYC \\
TOP2A & Proliferation-machinery & DNA topoisomerase & BRCA & 100.0\% & $<0.0001$ & Replicates MYC \\
TOP2A & Proliferation-machinery & DNA topoisomerase & STAD & 100.0\% & $<0.0001$ & Replicates MYC \\
RPS6KB1 & Proliferation-machinery & Ribosomal S6 kinase & BRCA & 89.6\% & $<0.0001$ & Replicates MYC \\
ERBB2 & Lineage-identity / RTK & RTK & BRCA & 18.0\% & 1.000 & Anti-concordant \\
ERBB2 & Lineage-identity / RTK & RTK & COAD & 14.3\% & 1.000 & Anti-concordant \\
ERBB2 & Lineage-identity / RTK & RTK & STAD & 93.6\% & $<0.0001$ & Concordant, inconsistent \\
SOX9 & Lineage-identity / RTK & Transcription factor & BRCA & 38.8\% & 1.000 & Anti-concordant \\
FOXA1 & Lineage-identity / RTK & Transcription factor & BRCA & 34.0\% & 1.000 & Anti-concordant \\
GATA3 & Lineage-identity / RTK & Transcription factor & BRCA & 0.0\% & 1.000 & Anti-concordant (complete) \\
ESR1 & Lineage-identity / RTK & Transcription factor & BRCA & 4.0\% & 1.000 & Anti-concordant, predicted \\
\bottomrule
\end{tabular}%
}
\renewcommand{\arraystretch}{1.0}
\vspace{2pt}
\footnotesize ``Replicates MYC'' refers to concordance rate only, independent of the baseline comparison in Section~\ref{sec:baseline_comparison}, which was run for E2F3 but not for the other genes in this table. The Type column reports each gene's established molecular function (e.g.\ NCBI Gene, UniProt) and is independent of, and not always identical to, CASCADE's own network-derived role classification (Table~\ref{tab:role_applied}); it is not a CASCADE output.
\end{table}

Candidate genes with strong, recurrent, cross-tissue amplification comparable to MYC are themselves rare -- lineage-restricted transcription factors such as KLF5, CTNNB1, and MYB lack sufficient amplified-sample counts in at least two of the three tested cancer types to run this test at all, which constrained the achievable panel size for this generalization check. We report the fifteen-gene, twenty-seven-combination comparison in Table~\ref{tab:other_genes} as evidence bearing on generalizability, not as a comprehensive panel-level statistic in the style of RegNetAgents' twelve-gene (BRCA) and thirteen-gene (COAD) panels, and discuss a candidate explanation for the observed split in Section~\ref{sec:discussion}.

Table~\ref{tab:other_genes}'s BH-FDR column applies Benjamini-Hochberg correction across all twenty-seven gene-cancer-type combinations tested in this generalization sweep (the twenty-four rows shown plus MYC's three cancer types from Table~\ref{tab:myc_main}): seventeen of twenty-seven combinations remain significant at $q<0.05$ and ten do not, with CCND1/STAD and CCND2/COAD as the two intermediate cases ($q=0.29$ and $q=0.47$ respectively). The resulting proliferation-machinery/lineage-identity split should still be read as a pattern we observed and report, not a pre-registered or formally tested hypothesis -- and, as the cross-cancer-type extension above shows, gene class alone is not the whole story; CCND2's outright failure alongside its validating paralogs, in both cancer types tested, is a further instance of that same point.

Table~\ref{tab:role_applied} applies CASCADE's gene-role classification (Table~\ref{tab:role_rules}) to every gene-cancer-type combination tested in the patient-data concordance experiments, reported descriptively as a record of what the classifier did, not as a variable systematically tested for association with outcome; with fifteen non-MYC genes across twenty-seven combinations (Section~\ref{sec:panel_selection}), any pattern here is an observation, not evidence. Role classification does not predict validation outcome. The sharpest single-gene example is ERBB2: classified as a master regulator in both BRCA and STAD, with no change in role label, yet anti-concordant in BRCA (18.0\%) and concordant in STAD (93.6\%) -- the same gene, the same role, opposite outcomes, driven by cancer-type context rather than anything CASCADE's classifier captures. The sharpest cross-gene example is CCND2 and CCND3: both classified as transcription factors in BRCA with comparable out-degree (34 and 48), the same paralog family, the same cancer type, and the same role label, yet CCND2 fails outright (38.0\%, anti-concordant) while CCND3 validates (76.0\%). Together, these are sufficient to rule out role classification as a strict, standalone explanation, independent of sample size -- though neither identifies what the actual explanation is.

\begin{table}[h]
\centering
\small
\caption{CASCADE gene-role classification (Table~\ref{tab:role_rules}) applied to genes tested in Sections~\ref{sec:myc_main} and~\ref{sec:gene_specificity}.}
\label{tab:role_applied}
\begin{tabular}{llcl}
\toprule
Gene & Cancer type & Out-degree & CASCADE role \\
\midrule
MYC & BRCA & 80 & Master regulator \\
MYC & COAD & 189 & Master regulator \\
MYC & STAD & 150 & Master regulator \\
E2F3 & BRCA & 91 & Master regulator \\
CCND1 & BRCA & 51 & Master regulator \\
CCND1 & STAD & 53 & Master regulator \\
CCND2 & BRCA & 34 & Transcription factor \\
CCND2 & COAD & 48 & Transcription factor \\
CCND3 & BRCA & 48 & Transcription factor \\
CCND3 & STAD & 102 & Master regulator \\
AURKA & BRCA & 87 & Master regulator \\
AURKA & COAD & 47 & Transcription factor \\
CCNE1 & BRCA & 89 & Master regulator \\
CCNE1 & STAD & 72 & Master regulator \\
MDM2 & BRCA & 63 & Master regulator \\
MDM2 & STAD & 132 & Master regulator \\
FOXM1 & BRCA & 330 & Master regulator \\
TOP2A & BRCA & 59 & Master regulator \\
TOP2A & STAD & 226 & Master regulator \\
RPS6KB1 & BRCA & 31 & Transcription factor \\
ERBB2 & BRCA & 109 & Master regulator \\
ERBB2 & COAD & 36 & Transcription factor \\
ERBB2 & STAD & 56 & Master regulator \\
SOX9 & BRCA & 37 & Transcription factor \\
FOXA1 & BRCA & 275 & Master regulator \\
GATA3 & BRCA & 251 & Master regulator \\
ESR1 & BRCA & 170 & Master regulator \\
\bottomrule
\end{tabular}
\end{table}

\FloatBarrier
\subsection{Agent tool-call grounding is reliable with a capable model, with three disclosed failure modes -- two resolved, one open}
\label{sec:agent_grounding_results}

\textbf{Overall accuracy.} Table~\ref{tab:agent_grounding} reports exact parameter-match accuracy across the 35-query benchmark (Section~\ref{sec:agent_grounding}), scored against each model's raw tool-call output. \texttt{llama3.1:8b}, CASCADE's own documented local-deployment default, reaches 71.4\% (25/35) exact match; \texttt{qwen2.5:72b}, tested as a secondary, larger-model comparison, reaches 85.7\% (30/35). The gap between them is not uniform across failure types, and each type has a distinct, disclosable explanation rather than being unexplained noise. Ollama's sampling is not seeded, so \texttt{llama3.1:8b}'s exact figure and specific failing queries can vary between runs; we report the one run for which we retained full per-query results throughout this section, the same run used to test the perturbation-type fix described below.

\begin{table}[H]
\centering
\small
\caption{Agent tool-call grounding accuracy by category, exact parameter match out of 5 queries per category, scored against each model's raw tool-call output.}
\label{tab:agent_grounding}
\begin{tabular}{lcc}
\toprule
Category & \texttt{llama3.1:8b} & \texttt{qwen2.5:72b} \\
\midrule
Baseline TCGA & 3/5 & 5/5 \\
Baseline immune & 5/5 & 5/5 \\
Gene alias$^\dagger$ & 1/5 & 2/5 \\
Cancer-type name & 4/5 & 5/5 \\
Perturbation phrasing & 4/5 & 5/5 \\
Implicit parameters & 4/5 & 3/5 \\
Multi-entity distractor & 4/5 & 5/5 \\
\midrule
\textbf{Overall} & \textbf{25/35 (71.4\%)} & \textbf{30/35 (85.7\%)} \\
\bottomrule
\end{tabular}
\vspace{2pt}
\footnotesize $^\dagger$Raw model output; scored instead against what CASCADE's server would do after its own \texttt{resolve\_alias()} step, Gene alias rises to 3/5 (\texttt{llama3.1:8b}) and 5/5 (\texttt{qwen2.5:72b}), and Overall rises to 27/35 (77.1\%) and 33/35 (94.3\%) respectively (Section~\ref{sec:agent_grounding_results}).

\vspace{2pt}
\footnotesize $^\ddagger$We additionally implemented and tested a fix for the ambiguous-query \texttt{perturbation\_type} default (Section~\ref{sec:agent_grounding_results}). It triggered on 0 of all 35 queries, for both models -- and both zeros share one root cause rather than being independent results: neither model ever left \texttt{perturbation\_type} unset, so the fix's trigger condition (a genuinely omitted field) never occurred anywhere in the benchmark. Consequently it flagged \emph{0/3 of the ambiguous queries it was designed to catch} (a failure to engage, not a null result) and, for the identical reason, \emph{0/32 of the remaining non-ambiguous queries} (no spurious clarifications, but this reflects the same never-engaged mechanism, not a validated regression check). The fix's own optional \texttt{query} parameter, needed for it to engage at all, was populated by \texttt{llama3.1:8b} in 23/35 calls overall (65.7\%, though never on the 3 ambiguous queries specifically) and by \texttt{qwen2.5:72b} in 0/35 calls.
\end{table}

\textbf{Network-parameter errors.} Network-parameter errors -- \texttt{tcga\_network} or \texttt{network\_source} left unset or set to an incorrect value -- occurred eight times with \texttt{llama3.1:8b} and zero times with \texttt{qwen2.5:72b} (Table~\ref{tab:agent_failures}): six on queries outside the Gene-alias category, discussed here, and two more on Gene-alias-category queries (HER3, p53), discussed separately below where they explain \texttt{llama3.1:8b}'s two remaining post-resolution failures. Of the six discussed here, three are true schema-adherence failures: a required \texttt{tcga\_network} value omitted entirely despite \texttt{network\_source} correctly set to \texttt{tcga}, which would cause CASCADE's real server to reject the call outright. The remaining three are wrong-value substitutions rather than omissions: two cancer-subtype disambiguation errors in different tissue pairs, and one \texttt{network\_source} substitution on an implicit-parameter query (``What does MYC do?'') routed to a TCGA network instead of the implied cell-type default -- the same query discussed further below as this benchmark's shared perturbation-type ambiguity failure. All eight errors are specific to the smaller model: \texttt{qwen2.5:72b} produced neither an omission nor a wrong-value network-parameter error anywhere in the benchmark.

\begin{table}[H]
\centering
\footnotesize
\caption{Individual failing queries underlying the three failure modes discussed in this section, by model, field, and value. Category names match Table~\ref{tab:agent_grounding}'s columns directly. A query with more than one field error (e.g.\ the HER3 query below, or ``What does MYC do?'' for \texttt{llama3.1:8b}, which recurs under both Network-parameter and Perturbation-type ambiguity errors) appears once per error, so row counts within a category can exceed Table~\ref{tab:agent_grounding}'s per-category error count, which counts distinct failing queries. ``Server-resolved'' marks gene-field mismatches that CASCADE's real \texttt{resolve\_alias()} step corrects (Section~\ref{sec:agent_grounding_results}); it is not applicable (N/A) to non-gene fields.}
\label{tab:agent_failures}
\setlength{\tabcolsep}{4pt}
\resizebox{\linewidth}{!}{%
\begin{tabular}{llllllc}
\toprule
Query & Category & Model & Field & Expected & Actual & Server-resolved \\
\midrule
\multicolumn{7}{l}{\textit{Network-parameter errors}} \\
EGFR knockdown, cervical cancer & Baseline TCGA & \texttt{llama3.1:8b} & \texttt{tcga\_network} & \texttt{cesc} & \texttt{ucec} & N/A \\
BRCA1 overexpression, ovarian cancer & Baseline TCGA & \texttt{llama3.1:8b} & \texttt{tcga\_network} & \texttt{ov} & (omitted) & N/A \\
SMAD4 knockdown, pancreatic cancer & Cancer-type name & \texttt{llama3.1:8b} & \texttt{tcga\_network} & \texttt{paad} & (omitted) & N/A \\
TP53 suppression, stomach cancer & Perturbation phrasing & \texttt{llama3.1:8b} & \texttt{tcga\_network} & \texttt{stad} & \texttt{paad} & N/A \\
``What does MYC do?'' & Implicit parameters & \texttt{llama3.1:8b} & \texttt{network\_source} & \texttt{cell\_type} & \texttt{tcga} & N/A \\
MYC+TP53 multi-entity, breast cancer & Multi-entity distractor & \texttt{llama3.1:8b} & \texttt{tcga\_network} & \texttt{brca} & (omitted) & N/A \\
\multicolumn{7}{l}{\textit{Gene-alias errors (raw output)}} \\
HER2 knockdown, breast cancer & Gene alias & \texttt{llama3.1:8b} & \texttt{gene} & \texttt{ERBB2} & \texttt{HER2} & Yes \\
HER2 knockdown, breast cancer & Gene alias & \texttt{qwen2.5:72b} & \texttt{gene} & \texttt{ERBB2} & \texttt{HER2} & Yes \\
PD-L1 knockdown, NK cells & Gene alias & \texttt{llama3.1:8b} & \texttt{gene} & \texttt{CD274} & \texttt{PD-L1} & Yes \\
PD-L1 knockdown, NK cells & Gene alias & \texttt{qwen2.5:72b} & \texttt{gene} & \texttt{CD274} & \texttt{PD-L1} & Yes \\
HER3 overexpression, stomach cancer & Gene alias & \texttt{llama3.1:8b} & \texttt{gene} & \texttt{ERBB3} & \texttt{HER3} & Yes \\
HER3 overexpression, stomach cancer & Gene alias & \texttt{llama3.1:8b} & \texttt{tcga\_network} & \texttt{stad} & \texttt{stomach} & N/A \\
HER3 overexpression, stomach cancer & Gene alias & \texttt{qwen2.5:72b} & \texttt{gene} & \texttt{ERBB3} & \texttt{HER3} & Yes \\
p53 overexpression, lung squamous carcinoma & Gene alias & \texttt{llama3.1:8b} & \texttt{tcga\_network} & \texttt{lusc} & \texttt{luad} & N/A \\
\multicolumn{7}{l}{\textit{Perturbation-type ambiguity errors}} \\
``What does MYC do?'' & Implicit parameters & \texttt{llama3.1:8b} & \texttt{perturbation\_type} & \texttt{knockdown} & \texttt{overexpression} & N/A \\
``What does MYC do?'' & Implicit parameters & \texttt{qwen2.5:72b} & \texttt{perturbation\_type} & \texttt{knockdown} & \texttt{overexpression} & N/A \\
``What genes are affected by GATA3?'' & Implicit parameters & \texttt{qwen2.5:72b} & \texttt{perturbation\_type} & \texttt{knockdown} & \texttt{overexpression} & N/A \\
\bottomrule
\end{tabular}%
}
\end{table}

\textbf{Gene-alias failures: raw output vs.\ server-resolved.} Gene-alias failures illustrate the gap between a model's raw output and what CASCADE's server actually does with it (Table~\ref{tab:agent_failures}). Scored against raw output, \texttt{llama3.1:8b} gets 1/5 gene-alias queries exactly right and \texttt{qwen2.5:72b} gets 2/5: both emit the literal informal name rather than the network's official symbol for three of the five queries, while both correctly produce the alias already in CASCADE's table without needing correction (p53$\to$\texttt{TP53}) and the alias resolvable via general reasoning rather than a table lookup (``the estrogen receptor''$\to$\texttt{ESR1}). Scored against what CASCADE's server would actually do -- running each model's raw \texttt{gene} value through the real \texttt{resolve\_alias()} function before comparing it -- every one of these three aliases resolves correctly for both models, since all three are covered by CASCADE's alias table (Section~\ref{sec:agent_grounding}), lifting the category to 3/5 and 5/5 respectively (Table~\ref{tab:agent_grounding}, note $\dagger$).

\texttt{llama3.1:8b}'s two remaining post-resolution failures are not gene-alias errors at all -- both are network-parameter errors on the same queries (Table~\ref{tab:agent_failures}). In other words, once CASCADE's own resolution step is accounted for, no gene name in this benchmark is left genuinely unresolved for either model: what remains is explained entirely by CASCADE's alias-table coverage (complete for every alias tested here) and, separately, by network-parameter errors unrelated to gene naming.

\textbf{A third failure mode: perturbation-type ambiguity.} A third failure mode -- defaulting to the wrong \texttt{perturbation\_type} on queries with no explicit directional cue -- affected both models, overlapping on one query and diverging on another (Table~\ref{tab:agent_failures}): both \texttt{llama3.1:8b} and \texttt{qwen2.5:72b} defaulted to \texttt{overexpression} instead of CASCADE's documented \texttt{knockdown} default on the same query (``What does MYC do?''), and \texttt{qwen2.5:72b} showed the identical pattern separately on a second query that \texttt{llama3.1:8b} answered correctly. That the pattern recurs across both models -- on a query they share and one specific to the larger, more capable model -- suggests this reflects a genuine mismatch between how such requests read naturally and what CASCADE's schema silently defaults to -- though with only three ambiguous queries tested, this is a suggestive pattern, not a robust estimate -- not an artifact of one query's specific wording or one model's limitations. Unlike the first two failure modes, this one is neither resolved by scale nor explained by a fixable gap in CASCADE's alias coverage, discussed further in Section~\ref{sec:limitations}.

\textbf{A targeted fix.} We designed and implemented a targeted, narrowly-scoped fix for this failure mode (Section~\ref{sec:agent_grounding}): a check, added to \texttt{cascade\_langgraph\_mcp\_server.py}'s \texttt{comprehensive\_\allowbreak perturbation\_\allowbreak analysis} handler, that scans the caller's original request text for directional cues whenever \texttt{perturbation\_type} is omitted from the tool call, returning a \texttt{clarification\_needed} response instead of silently defaulting when no single direction is detected; the check never overrides an explicitly-supplied value, by design, so it cannot introduce a regression on any query where a model already states a \texttt{perturbation\_type}.

\textbf{Testing reveals a different problem.} Testing this fix against the same three ambiguous-directional queries revealed something more fundamental than whether the fix worked: across all six ambiguous-query calls tested (three queries $\times$ two models), neither model ever left \texttt{perturbation\_type} unset -- both always supplied a concrete value, correct or not (Table~\ref{tab:agent_grounding}, note $\ddagger$) -- so the check's trigger condition, a genuinely omitted field, never actually occurred. This reframes the original failure: it is not an omission-and-silent-default problem but a confident-wrong-guess problem. Instruction-tuned models, at least the two tested here, appear to reliably populate enum-typed schema fields even under genuine input ambiguity, rather than signaling uncertainty by leaving a field unset. A server-side check gated on a missing field is consequently inert against this failure mode -- correctly scoped to the condition we set out to fix, but addressing a condition that, empirically, does not arise with either model tested. Resolving the underlying failure would require a fix that engages even when a value is present -- for example, validating a supplied value against the query's actual textual support, or requiring an explicit ``unspecified'' sentinel value rather than allowing silent selection among enum options -- which is outside the scope of what we implemented and tested here.

\FloatBarrier
\section{Discussion}
\label{sec:discussion}

The central claim this paper supports about CASCADE specifically is narrow and specific: for MYC, across three cancer types, under a PAM50 subtype control, in a fully independent patient cohort, and stable across a fourfold range of panel size, CASCADE's downstream knockdown-propagation predictions show real, statistically robust, biologically coherent concordance with patient tumor data. This is a different -- and, for the specific claim being validated, more direct -- form of evidence than driver-gene enrichment: it tests whether a \emph{specific predicted direction of change} matches what happens in actual disease, rather than whether a predicted gene is plausible in general.

That concordance result establishes CASCADE's predictions are trustworthy once correctly invoked; Section~\ref{sec:agent_grounding_results} addresses a distinct, upstream claim -- whether an agent calling CASCADE via MCP correctly constructs that invocation in the first place, using a locally-hosted model with no dependency on a proprietary API. This claim is more qualified than the biological one: it holds only partially for CASCADE's own smaller documented default, and one identified failure mode -- perturbation-type ambiguity -- persists regardless of model size (Section~\ref{sec:agent_grounding_results}).

This result complements RegNetAgents' validation of the converse analytical direction (upstream regulator-candidate identification): where RegNetAgents tests \emph{membership} via driver-gene enrichment, CASCADE's task makes a \emph{directional} claim that only patient-data concordance tests directly. The biologically coherent, patient-data-confirmed MYC targets in Table~\ref{tab:myc_worked} (NOP14, RIOK1, WDR43, KAT2A, and the like) illustrate why: they are ribosome-biogenesis components, none of them listed as an IntOGen cancer driver \citep{intogen}, yet their predicted direction of change is independently confirmed correct. The eligibility-passing genes that failed to validate in Section~\ref{sec:gene_specificity} (GATA3, FOXA1, SOX9, ESR1) serve as a retrospective specificity check: each satisfies the identical structural and statistical-power prerequisites as the validating genes, yet produces null-to-anti-concordant results.

No single mechanism yet explains why MYC, E2F3, AURKA, CCNE1, FOXM1, TOP2A, RPS6KB1, and CCND3 validate consistently wherever tested (68--100\%); why CCND1 and MDM2 validate in BRCA but not STAD, while ERBB2 shows the reverse pattern -- anti-concordant in BRCA and COAD, concordant in STAD; why CCND3 validates in both cancer types tested but its paralog CCND2 does not, in either, despite an identical eligibility profile; or why SOX9, FOXA1, GATA3, and ESR1 fail outright. A transcription-factor/non-transcription-factor split does not explain it: none of CCND1, CCND3, AURKA, CCNE1, MDM2, TOP2A, or RPS6KB1 is a transcription factor by molecular function, yet all validate in at least one cancer type, while SOX9, FOXA1, GATA3, and ESR1 are genuine transcription factors that all fail -- and CCND2, also not a transcription factor, fails too, so transcription-factor status predicts neither outcome.

We propose, as a hedged post-hoc hypothesis rather than an established mechanism, a second explanation that fits the data more closely, though it remains speculative and untested by direct mechanistic evidence. MYC, E2F3, CCND1, CCND3, AURKA, CCNE1, MDM2, FOXM1, TOP2A, and RPS6KB1 all sit within or adjacent to core proliferation-machinery, whose output is plausibly expected to scale with regulator dosage across many tissue contexts. GATA3, FOXA1, SOX9, and ESR1, by contrast, are lineage-identity transcription factors that establish cell state via chromatin-level mechanisms that may not respond linearly to added gene copies. CCND2 does not fit either side: it shares CCND1 and CCND3's exact molecular function, yet fails outright in every cancer type tested (38.0\% in BRCA, 54.2\% in COAD) where its paralogs validate, so some factor beyond this distinction must separate it from its own paralogs.

GATA3's result is the sharpest failure in the panel (0 of 50 predicted targets concordant), and not an artifact of the embedding blend or unreliable ARACNe annotation -- its highest-confidence target edges (ESR1, MLPH, XBP1) are correctly annotated as activating (Section~\ref{sec:gene_specificity}). One candidate explanation, untested here, is that copy-number amplification is simply the wrong dosage proxy for GATA3, whose oncogenic role in breast cancer is driven predominantly by point mutation rather than amplification.

The cross-cancer-type extension (Section~\ref{sec:gene_specificity}) adds a further wrinkle: proliferation-machinery membership predicts validation in \emph{at least one} cancer type well (nine of ten non-MYC genes validate somewhere), but guarantees neither validation in every eligible cancer type (CCND1, MDM2) nor, as CCND2 shows, validation in any cancer type at all despite sharing its validating paralogs' eligibility profile. Candidate factors -- tumor purity, stromal contamination, amplicon co-selection -- remain untested here. Proliferation-machinery membership is necessary but not sufficient for validation, and ERBB2's earlier cancer-type inconsistency (Section~\ref{sec:gene_specificity}) turns out to be an early hint of this more general pattern.

The baseline comparison (Section~\ref{sec:baseline_comparison}) tempers this picture: neither the MYC-identity nor the E2F-identity comparison distinguishes CASCADE's accuracy from its curated baseline at conventional significance, before or after multiple-testing correction, though CASCADE reaches this from zero gene-specific curation, unlike the curated lists' dedicated perturbation datasets and formal Hallmark curation procedure \citep{liberzon2015}. CASCADE's gene-specific direction-calling is not redundant with a naive guess, however, as the ablation reported above shows.\footnote{This pattern is not unique to CASCADE: GEARS' own benchmarking includes a baseline adapted from CellOracle \citep{celloracle} that infers a gene regulatory network and linearly propagates perturbation signal along it -- structurally similar to CASCADE's own network-propagation approach -- and a recent broader benchmark reports that a naive average-perturbation-effect baseline (the ``perturbed mean'') often matches or exceeds GEARS, scGPT, and CPA \citep{systema}. A simple baseline matching or exceeding a more sophisticated model recurs across the perturbation-prediction literature generally, and is not an isolated weakness of the comparison in Section~\ref{sec:baseline_comparison}.}

\subsection{Limitations}
\label{sec:limitations}

The TCGA ARACNe networks CASCADE uses are population-averaged per cancer type and do not capture tumor subclone heterogeneity or single-patient regulatory state. Copy-number amplification is a proxy for gene dosage, not a direct experimental perturbation, and the copy-number/activity relationship is not perfectly linear for every gene -- partially, not fully, mitigated by the PAM50 control and METABRIC replication.

Our generalization panel covers fifteen genes in BRCA (sixteen with MYC) against RegNetAgents' twelve, and three genes in COAD (four with MYC) against its thirteen; RegNetAgents did not test STAD. Panel size in this generalization sweep was constrained by the rarity of genes with MYC-comparable cross-tissue amplification. We attempted a direct experimental test of the lineage-identity hypothesis using LINCS L1000 shRNA data (GSE106127, MCF7) for GATA3, FOXA1, SOX9, and ESR1, but target coverage was insufficient: only 2--5 of each gene's top-50 predicted targets are on the 978-gene L1000 panel, and SOX9 was absent from the shRNA reagent library entirely (\texttt{scripts/experiment5\_lincs\_coverage\_check.py}).

The baseline comparison (Section~\ref{sec:baseline_comparison}) is similarly limited -- two curated MSigDB gene lists (\texttt{MYC\_TARGETS\_V1}, \texttt{E2F\_TARGETS}; Appendix~\ref{app:baseline}), the latter tested only in BRCA -- too few to characterize CASCADE's accuracy against curated annotation more broadly. A head-to-head comparison against CellOracle \citep{celloracle}, the closest existing network-propagation approach, remains open future work. Because perturbation prediction is CASCADE's central function rather than a secondary capability, this gap matters more here than it would for a narrower claim; the patient-data concordance results above test that central function against real outcomes directly, but they do not establish CASCADE's accuracy relative to the nearest comparable tool.

A related concern is that the sequence of gene selections in this paper could reflect post-hoc pattern-fitting rather than a genuine effect: some later focal genes were chosen after a working hypothesis predicted their outcome (Section~\ref{sec:gene_specificity}). No eligible gene was dropped along the way, and each gene's selection provenance -- blind screen, hypothesis-motivated addition, or hypothesis-falsifying test case -- is disclosed at the point it is introduced. This does not establish the hypothesis is correct, and the sequence includes a gene (ESR1) selected specifically to falsify the pattern, not only genes selected to confirm it.

The agent tool-call grounding benchmark (Section~\ref{sec:agent_grounding_results}) is similarly bounded: 35 queries against two local models characterizes distinct failure modes but not a precise, generalizable accuracy estimate, and we did not test proprietary frontier models, which plausibly perform better on schema-adherence. CASCADE's gene-alias table (\texttt{tools/gene\_id\_mapper.py}) covers a small, fixed set of common informal gene names; the server-resolved figures (77.1\%, 94.3\%) should not be read as evidence of generalization to aliases beyond that coverage. The ambiguous-query perturbation-type fix (Section~\ref{sec:agent_grounding_results}) remains open: it is correctly scoped but inert against the confident-wrong-guess failure actually observed, and a fix that validates a supplied value against the query's textual support remains future work. This benchmark also evaluates a single agentic decision point in isolation, not sustained multi-turn behavior or recovery from a failed tool call -- a fuller behavioral evaluation is future work.

\section{Funding}

No funding was received for this work.

\section{Competing Interests}

The author is the founder of Bird AI Solutions. This work was conducted independently and received no funding.

\section{Acknowledgements}

CASCADE's embedding-enhanced propagation path, described in Section~\ref{sec:architecture} and used throughout this paper's experiments, uses pre-trained gene embeddings from the GREmLN model developed by the Chan Zuckerberg Initiative AI team \citep{gremln}. The GREmLN cell-type network data were obtained from the GREmLN team's pre-computed ARACNe networks \citep{gremln}. We acknowledge the TCGA Research Network, the ARACNe network compilation \citep{aracnenetworks,aracnezenodo}, the cBioPortal team, the METABRIC consortium, and the IntOGen team for the public data and APIs this work depends on. The results shown here are in whole or part based upon data generated by the TCGA Research Network: \url{https://www.cancer.gov/tcga}.

\section{AI Usage Disclosure}

Claude Code (Anthropic) was used to assist with software implementation, statistical analysis, and paper drafting. The author reviewed and approved all analyses and text and takes full responsibility for the methods and conclusions.

\section{Data and Code Availability}

CASCADE is available at \url{https://github.com/jab57/CASCADE} under the MIT license, archived at Zenodo (the v1.5.0 release used for this paper: \url{https://doi.org/10.5281/zenodo.23248601}; all versions: \url{https://doi.org/10.5281/zenodo.18883012}). GREmLN project ARACNe network files were obtained from the GREmLN Quickstart Tutorial (Chan Zuckerberg Initiative Virtual Cells Platform; \url{https://virtualcellmodels.cziscience.com/quickstart/gremln-quickstart}; model code: \url{https://github.com/czi-ai/GREmLN}) under the Virtual Cells Platform Terms of Use (scientific research use) and are bundled with CASCADE for convenience; they derive from CELLxGENE Census data \citep{cellxgene} (CC BY 4.0, \url{https://creativecommons.org/licenses/by/4.0/}). No separate data DOI exists for these files. The TCGA ARACNe networks are those of Bioconductor \texttt{aracne.networks}, whose network files the package authors publish on Zenodo (doi:10.5281/zenodo.22918956; CC BY-NC-ND 4.0, \url{https://creativecommons.org/licenses/by-nc-nd/4.0/}); CASCADE converts them locally to CSV format and does not redistribute them. A script in the repository (\texttt{scripts/extract\_tcga\_networks.py}) installs them from that record. Network-derived values (the predicted targets in Table~\ref{tab:myc_worked} and the network out-degrees in Table~\ref{tab:role_applied}) remain subject to the CC BY-NC-ND 4.0 license of the TCGA network data and are not covered by this article's license. The repository README documents CASCADE's full tool set and usage examples beyond the propagation-and-embedding component validated in this paper. Validation scripts for this paper and their full results are included in the repository under \texttt{scripts/} and \texttt{supplementary/} respectively:
\begin{itemize}
\item \texttt{scripts/experiment4\_tcga\_myc\_concordance.py} -- primary patient-data concordance test, parametrized by cancer type and focal gene
\item \texttt{scripts/experiment4\_pam50\_control.py} -- PAM50 subtype-controlled subgroup test
\item \texttt{scripts/experiment4\_metabric\_replication.py} -- independent-cohort replication
\item \texttt{scripts/experiment4\_hallmark\_baseline.py} -- curated, identity-matched gene-set baseline comparison (Section~\ref{sec:baseline_comparison})
\item \texttt{scripts/experiment4\_baseline\_comparison\_stats.py} -- combines CASCADE's and each baseline's concordance counts into the Fisher's-exact and BH-FDR statistics reported in Table~\ref{tab:baseline_comparison} (Section~\ref{sec:baseline_comparison})
\item \texttt{scripts/experiment4\_generalization\_bh\_fdr.py} -- Benjamini-Hochberg correction across the twenty-seven generalization gene-cancer-type combinations (Section~\ref{sec:gene_specificity}, Table~\ref{tab:other_genes})
\item \texttt{scripts/\allowbreak experiment4\_\allowbreak direction\_\allowbreak split\_\allowbreak check.py} -- checks whether CASCADE's own predicted-target panels are themselves direction-skewed (Section~\ref{sec:baseline_comparison})
\item \texttt{scripts/experiment5\_lincs\_coverage\_check.py} -- checks LINCS L1000 shRNA target-coverage feasibility for a direct experimental test of the lineage-identity hypothesis (Section~\ref{sec:limitations})
\item \texttt{scripts/experiment6\_agent\_tool\_grounding.py} -- agent tool-call grounding benchmark against CASCADE's real MCP tool schema (Section~\ref{sec:agent_grounding})
\end{itemize}
Patient expression and copy-number data were accessed via the public cBioPortal REST API (\url{https://www.cbioportal.org/api}); cancer driver annotations from the IntOGen compendium \citep{intogen} (release 2024.09.20, CC0 1.0, \url{https://creativecommons.org/publicdomain/zero/1.0/}; \url{https://www.intogen.org}).

\appendix
\section{Full Baseline Comparison Results}
\label{app:baseline}

Section~\ref{sec:baseline_comparison} describes this comparison's methodology and interpretation; this appendix reports the full numeric results. The two MSigDB gene sets used are \texttt{MYC\_TARGETS\_V1} (tested against all three cancer types) and \texttt{E2F\_TARGETS} (tested in BRCA only, against E2F3) \citep{liberzon2015}, each compared against CASCADE's own $N{=}50$ panel using every resolvable gene in the set (188--197 genes depending on set and cancer type). Table~\ref{tab:baseline_comparison} reports all four resulting comparisons; the comparison is not exhaustive -- only two curated gene sets were tested, and E2F-identity only in BRCA -- so it should be read as a partial check, not a complete characterization.

\begin{table}[h]
\centering
\footnotesize
\setlength{\tabcolsep}{4pt}
\caption{Baseline comparison: CASCADE's predictions vs.\ curated, identity-matched non-CASCADE gene-set baselines. Fisher's exact $p$ (two-sided) tests whether CASCADE's concordance count and the baseline's concordance count, each out of their own respective sample size (CASCADE $N{=}50$, or $N{=}49$ in STAD; baseline $N{=}188$--$197$), differ significantly. BH-FDR $q$ applies Benjamini-Hochberg correction across all four comparisons in this table.}
\label{tab:baseline_comparison}
\resizebox{\linewidth}{!}{%
\begin{tabular}{llcccc}
\toprule
Comparison & Cancer type & CASCADE & Baseline (full-set) & Fisher's exact $p$ & BH-FDR $q$ \\
\midrule
MYC-identity (MYC\_TARGETS\_V1) & BRCA & 90.0\% & 94.2\% & 0.336 & 0.767 \\
MYC-identity (MYC\_TARGETS\_V1) & COAD & 72.0\% & 69.6\% & 0.863 & 0.863 \\
MYC-identity (MYC\_TARGETS\_V1) & STAD & 85.7\% & 86.9\% & 0.816 & 0.863 \\
E2F-identity (E2F\_TARGETS), vs.\ E2F3 & BRCA & 96.0\% & 90.7\% & 0.384 & 0.767 \\
\bottomrule
\end{tabular}%
}
\vspace{2pt}
\footnotesize None of the four comparisons is statistically significant, before or after Benjamini-Hochberg correction.
\end{table}

CASCADE's own predicted-target panels are themselves not close to a uniform ``down'' guess: full-panel down/up splits range from a near-even 47\%\slash 53\% (MYC/STAD) to 78\%\slash 22\% (E2F3/BRCA); the MYC/BRCA panel underlying Table~\ref{tab:myc_worked} itself is 74.0\%\slash 26.0\% (37/50 down), markedly less skewed than the 86.7\% ``down'' share visible in that table's top-15-by-magnitude slice. Forcing every predicted ``up'' gene to ``down'' -- removing CASCADE's direction information while keeping its gene selection unchanged -- drops concordance by 14--29 percentage points in every panel (MYC/BRCA: 90.0\%$\to$68.0\%; MYC/COAD: 72.0\%$\to$54.0\%; MYC/STAD: 85.7\%$\to$57.1\%; E2F3/BRCA: 96.0\%$\to$82.0\%). Considering only the subset of genes CASCADE predicts will go up (rather than down) upon knockdown, those up-predictions are themselves correct 76.5--92.3\% of the time across panels -- higher than the down-predictions in the same panel for MYC/BRCA (92.3\% vs.\ 89.2\%) and MYC/COAD (76.5\% vs.\ 69.7\%), lower for MYC/STAD (76.9\% vs.\ 95.7\%) and E2F3/BRCA (81.8\% vs.\ 100.0\%), with no consistent direction across panels but never close to the 50\% chance level. This rules out the possibility that CASCADE's minority ``up'' calls are just noise being carried by the accuracy of its more common ``down'' calls.

\FloatBarrier
\bibliographystyle{plainnat}

\end{document}